\documentclass{llncs}

\usepackage{llncsdoc}
\usepackage[T1]{fontenc}
\usepackage{cite}

\usepackage[utf8]{inputenc}

\usepackage{times}

\usepackage{amssymb}
\usepackage{amsmath}
\usepackage{xspace}
\usepackage{array}
\usepackage[pdftex]{graphicx}

\usepackage{color}

\usepackage{tabularx}

\newcolumntype{C}{>{\centering}X}
\newcommand{\nl}{\tabularnewline}

\title{A Model for Safety Case Confidence Assessment}

\author{J\'{e}r\'{e}mie Guiochet\inst{1,2} \and Quynh Anh Do Hoang\inst{1,2} \and Mohamed Kaaniche\inst{1,2}}
\institute{LAAS-CNRS, 7 avenue du colonel Roche, 31031 Toulouse, France \and Universit\'{e} de Toulouse, France \\
\email{\{guiochet, qdohoang, kaaniche\}@laas.fr}
}

\begin{document}
\maketitle
\thispagestyle{empty}
\pagestyle{empty}

\begin{abstract}
Building a safety case is a common approach to make expert judgement explicit about safety of a system. The issue of confidence in such argumentation is still an open research field. Providing quantitative estimation of confidence is an interesting approach to manage complexity of arguments. This paper explores the main current approaches, and proposes a new model for quantitative confidence estimation based on Belief Theory for its definition, and on Bayesian Belief Networks for its propagation in safety case networks. 


\end{abstract}


\section{Introduction}
Safety cases are used in several critical industrial sectors to justify safety of installations and operations. As defined in the standard \cite{UKDef3}: "a Safety Case is a structured argument, supported by a body of evidence, that provides a compelling, comprehensible and valid case that a system is safe for a given application in a given environment". An important research work has also been initiated to deliver guidelines to document safety cases. An initial work  developed at York University \cite{KEL98}, based on an adaptation of Toulmin argumentation model \cite{TOU58}, led to the proposal of the Goal Structuring Notation (GSN). Other proposals such as CAE for Claims-Argument-Evidence \cite{CAE} and KAOS (Knowledge Acquisition and autOmated Specification) \cite{DAR93}, but they did not reach the maturity of GSN \cite{GSN11}. The Object Management Group (OMG) has also delivered a metamodel for the argumentation approach \cite{OMGARM}. The goal of these approaches is to make more explicit the supporting arguments for a top-level claim.

\begin{figure}
	\center
	\includegraphics[width=10cm]{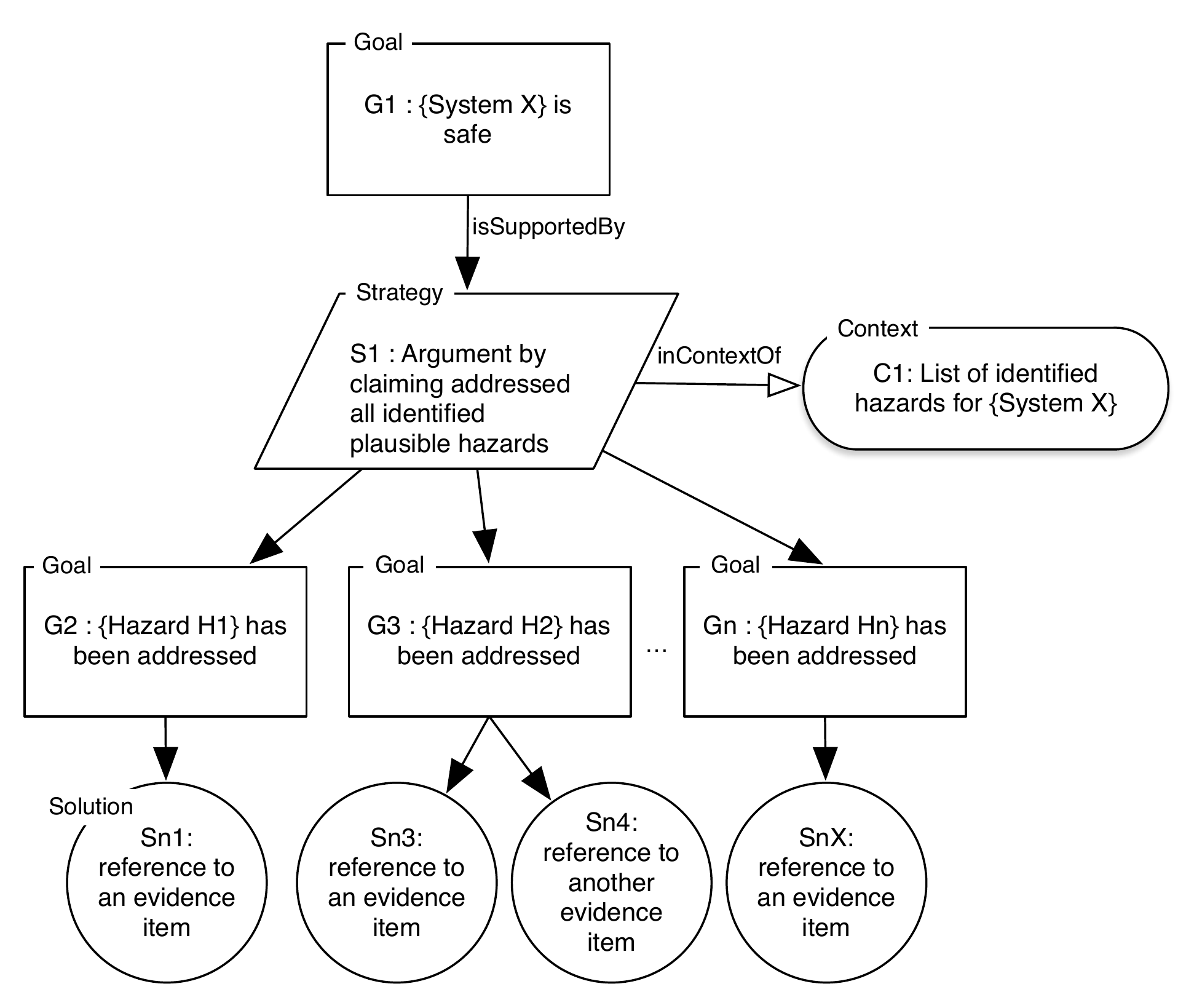}
	\caption{GSN example adapted from Hazard Avoidance Pattern \cite{KEL97} }
	\label{fig:gsn-example}
	\vspace{-0.4cm}
\end{figure}

Given a claim and a supporting argument, an important and growing issue is to understand how much confidence one could have in the claim and how the different arguments contribute to such confidence. For instance, let us consider the classical example of the claim "\{System X\} is safe", supported by the evidence that all specific hazards have been eliminated as presented in Figure~\ref{fig:gsn-example}. 
Main concepts of GSN are presented here: goals present claims forming part of the argument; Strategies describe the nature of inferences that exist between a goal and its supporting sub-goal(s); Solutions present a reference to an evidence item (results of a fault tree analysis for instance); Contexts present contextual artifacts (they could be a reference to contextual information, or statements). Other elements are used in GSN but not presented here as our proposal focuses on these main components of GSN. Each element of such an argument may be subject of uncertainties, such as "do all the hazards have been identified?" or "is the treatment of hazard $n$ effective?". Moreover, considering that argument structures tend to grow excessively, it may become too complex for third parties to analyse the argument. Therefore, appropriate methods to assess confidence in the argument structures and supporting evidence are required. Three main challenges are of particular interest: how confidence could be formally defined, how confidence could be quantitatively estimated, and how confidence in argument leaves could be propagated to assess the impact on the main claim confidence.

In this paper we mainly address the first and third issues by introducing a new method for defining and propagating a quantitative estimation of confidence of a safety case.   After presenting related work in Section~\ref{sec:relatedwork}, we introduce our definition of confidence based on belief theory in Section~\ref{sec:confdef}. This definition is used in Section~\ref{sec:propagconf} where details about confidence propagation are given. Finally, in the conclusion we will discuss about first results and open issues in this area. 
	\vspace{-0.4cm}


\vspace{-0.3cm}
\section{Related Work}

\label{sec:relatedwork} 

The issue of confidence in argument structures has already been addressed by several works, with slightly different objectives and scopes. Table~\ref{fig:biblioMethod} presents a common framework to analyze some relevant related work considering the following dimensions:
\begin{itemize}
\item Argument modelling: construction of the "case" which may be based on GSN or other notations
\item Argument uncertainties identification: uncertainties in inferences and arguments elements are identified
\item Confidence modelling: construction of a confidence case, with explicit representation of dependencies between the uncertainties 
\item Confidence estimation: theoretical framework for quantitative estimation of the confidence
\item Decision support: provide support based on the quantitative estimation in order to make a decision for the acceptability of the argument, or its improvement. 
\end{itemize}
\vspace{-0.7cm}
\renewcommand{\arraystretch}{1.2} 
\begin{table}
\begin{center}
\caption{Different approaches for managing confidence in safety case}
\scriptsize 
\usefont{T1}{phv}{l}{n}
 \begin{tabularx}{\textwidth}{|m{1cm}|C|C|C|C|C|} 
\cline{2-6} 
	\multicolumn{1}{c|}{}
		     & \textbf{Argument modelling} & \textbf{Argument uncertainties identification }& \textbf{Confidence modelling} & \textbf{Confidence estimation} & \textbf{Decision support} \nl \cline{2-6}		\hline
    \cite{LIT07} & &  & Bayesian network & Probability law &  \nl \hline
     \cite{Zhao12}  & Argumentation Metamodel (ARM) based case  & Based on Toulmin model & Bayesian network (with Hitchcock criteria) & Probability law (with basic 
logical gates) &  \nl \hline
     \cite{DEN11}  & GSN &  & Bayesian network & Probability law and tool support with AgenaRisk &  \nl \hline
     \cite{CYR11}  & Trust case based on Toulmin model &  &  & Dempster-Shafer Theory & Decision level associated to confidence level \nl \hline
     \cite{ANA13}  & GSN & &  & Dempster-Shafer Theory & Decision based on the confidence value \nl \hline
     \cite{ANA12}  & GSN & Common Characteristic Map (CCM) & Confidence case based on GSN &  &  \nl \hline
     \cite{GOO12}  & GSN & Based on Assurance Claim Points (ACP) & Confidence case in GSN & Baconian probability &  \nl \hline
    \cite{HAW11}  & GSN & Based on Assurance Claim Points (ACP) & Assurance case in GSN &  &  \nl \hline

\end{tabularx}
	\normalsize
	\normalfont

\label{fig:biblioMethod}
\end{center}
\vspace{-1cm}
\end{table}

\paragraph{Qualitative Approaches} 
~\\
~\\
In \cite{HAW11}, the inventors of GSN  address the confidence issue, by proposing to split a traditional safety case in two pieces. The first is the safety argument, showing all evidences, and the second is a confidence argument that addresses confidence in evidences, contexts, and individual inferences.  This confidence argument is also represented with GSN. It starts by adding to the safety case some possible uncertainty sources, which are called Assurance Claim Points (ACP), that are attached to inferences (the arrows connecting claims), contexts (explanatory information), or solutions. Then, for each ACP, an argumentation mainly focuses on demonstrating that the ACP is trustworthy and appropriate, which is built using GSN. 
Another proposal \cite{ANA12}, is based on the ACP but only focuses on Context and Solution elements. The authors propose to use a map (Common Characteristic Map) as a check list to identify sources of uncertainties, with recursive dependencies. For instance, if a safety case includes a solution which is a "Process result", they propose the generic uncertainties related to  "the use of a language", "the use of a tool", "the use of a mechanism", "the involved artifacts", etc. All those characteristics are then refined, with possible recursive dependencies. 

The proposed approach in \cite{GOO13} is quite similar, adapting the defeater concept from Defeasible Reasoning theory introduced by \cite{POL08}. These defeaters that could be compared to previous ACP, or weaknesses in the argumentation, are then analyzed to be reduced one by one.

Both previous proposals focus on the identification of the weaknesses in an argumentation, and present methods for a well structured approach. Nevertheless, such approaches may lead to complex confidence cases. Although controversial, we believe that quantitative estimation approaches may help to analyze the safety case confidence. For instance, it can support sensitivity analyses to identify the weak elements of an argumentation.

\paragraph{Quantitative Approaches} 
~\\
~\\
This group of approaches tries to apply mathematical formalism to capture lack of confidence in argument elements. Apart from some proposals based on simple mathematical models as in \cite{GOO12} where the number of uncertainties is estimated, two main ways of approaching the problem can be identified:
\begin{itemize}
\item Bayesian Belief Networks (BBNs): in this case the uncertainty is interpreted as a probability. BBNs are then applied to deduce the confidence in a goal from credibility of its backing arguments. Some authors directly use BBNs for modeling arguments and confidence. For instance, in \cite{HOB12}, they only use BBNs and commercial tools to calculate "trustworthy", which is actually a conditional probability. With a similar approach, authors of \cite{LIT07} particularly focus on the diversification in argumentation, calculating how a "multilegged" argument (a claim is supported by two evidences) impacts the probability (interpreted as a confidence level) of achieving the main claim. However, they directly use BBNs, without any safety case. On the contrary, \cite{Zhao12} propose to apply to each claim of a Toulmin model argument, a Bayesien network pattern showing relationships between uncertainties in the argumentation based on Hitchock criteria \cite{HIT05}. However, confidence propagation is not clearly analyzed and justified. In \cite{DEN11}, the authors present an interesting approach to build a BBN from the safety case, and use the work of \cite{FEN12}, to define a distribution of confidence for each argument element, but they do not propose transformation rules between safety case in GSN and the confidence BBN. The confidence propagation formulas are also not justified.   
\item Dempster–Shafer (D-S) theory of evidence. These approaches are based on the belief theory developed by P. Dempster in 1967, and extended by G. Shafer in 1976. A common justification for its use, is that probability theory does not make difference between epistemic and aleatory uncertainties \cite{AGU13}. In the D-S approach, belief, disbelief and  epistemic uncertainty are explicitly quantified. An important work by \cite{CYR11} is based on this theory. The authors, propose to build "Trust cases" based on Toulmin concepts, and to directly associate levels for belief and uncertainties, linked with a decision to accept or not an argument element. In this case, they do not build a confidence case, but directly propose a method and a tool for decision support. As presented later, they do not explicitly take into account confidence in the inferences of the argument. Authors of \cite{ANA13}, directly reuse the previous work, with a limited version, only considering that for each argument element it exists a level for "sufficiency". 
\end{itemize}

In summary, defining and measuring confidence in assurance claims is an important and open issue. A framework for determining confidence is needed, and this paper presents some initial steps to fulfill this objective.

\vspace{-0.3cm}
\section{Proposed Approach Overview}

\begin{figure}
\centering
\includegraphics[width=10cm]{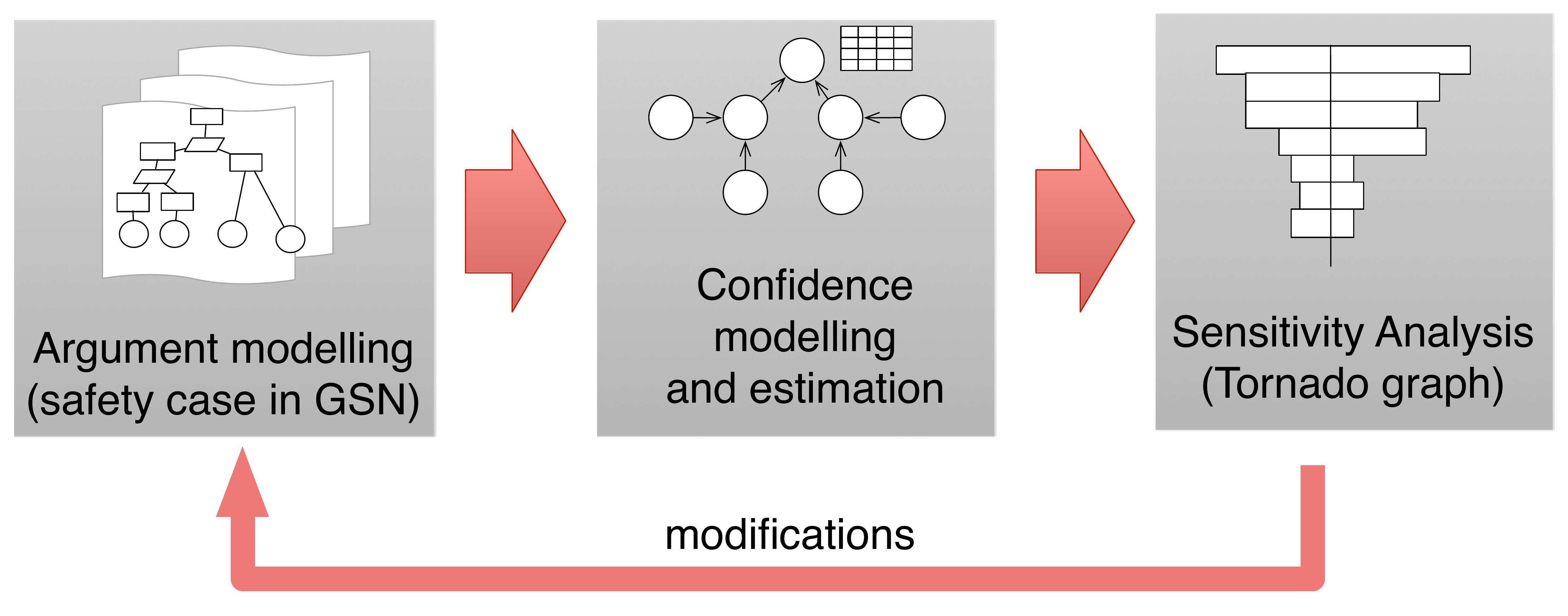}
\caption{Overview of the proposed method}
\label{fig:methodegenerale1}
\vspace{-0.8cm}
\end{figure}

Our objective is to propose a method to identify weaknesses in safety case, in order to improve it. Referring to Table~\ref{fig:biblioMethod}, our contribution focuses on the following steps presented Figure~\ref{fig:methodegenerale1}:
\begin{itemize}
\item Argument modelling: the safety case is built using GSN
\item Confidence modelling: we propose to annotate the GSN models and transform them into a confidence network
\item Confidence estimation: confidence in the network leaves are estimated and propagation formulas are used
\item Sensitivity analysis: impact of confidence variations is analyzed to identify weaknesses of the safety case. 
\end{itemize}


\vspace{-0.3cm}
\section{Measuring Confidence}
\label{sec:confdef} 
\begin{figure}
\centering
\includegraphics[width=6cm]{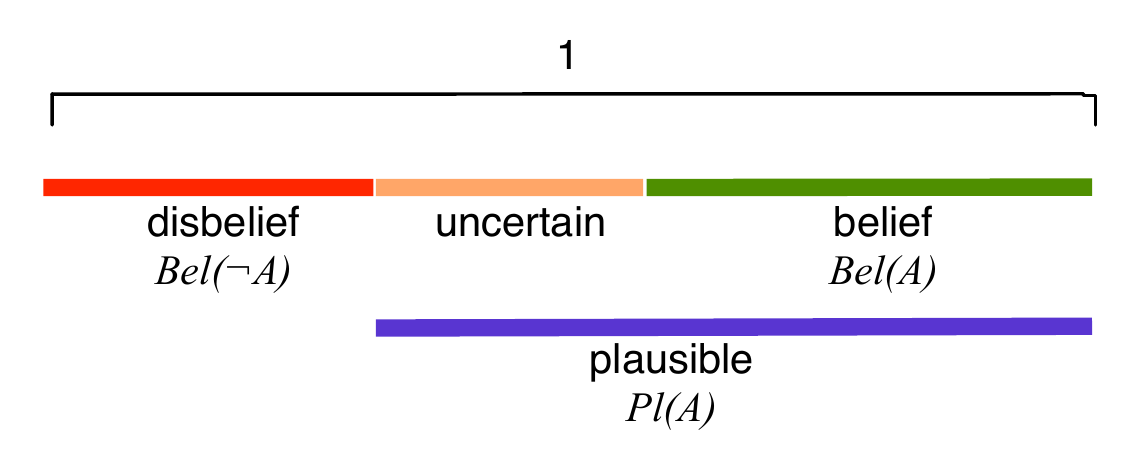}
\caption{D-S theory concepts, with a Boolean set}
\label{fig:ds1}
\vspace{-0.5cm}
\end{figure}

Confidence may be used as a common concept for different theories, including probability, and D-S. As in \cite{CYR11, ANA12}, we define confidence using the D-S approach. In this theory, a belief function is defined from the powerset $\mathcal{P}(\Omega)$ of possible events into $[0;1]$. For instance, let $\omega$ be the state of an indicator light that can have two values $on$ and $off$, then $\Omega=\{on, off\}$ and  $\mathcal{P}(\Omega)=\{\{on\},\{off\},\{on, off\},\varnothing \}$. In this example the belief function $Bel$, is defined as the mass $m$ of belief such as $Bel(\{on\})$ represents the credibility of the light to be $ON$. As an example, a possible estimation would be $Bel(\{on\})=m(\{on\})=0.2$, $Bel(\{off\})=m(\{off\})=0.5$ et $m(\{on, off\})=m(\Omega)=0.3$. When events are Boolean, like in this example, we can sum-up the D-S concepts with the Figure~\ref{fig:ds1} (Plausibility is another D-S concept which will not be included in this paper).

We will consider in a safety case that all elements leaves are observed, and that they cannot be false. Hence, for an element $A$, $Bel(\overline{A})=0$. This led us to define confidence and uncertainty as the belief functions:
\begin{equation}
\left\{
\begin{array}{l}
  m(A)=Bel(A)=g(A) \in [0,1] ~~:~~~confidence \\
  m(A, \overline{A}) =1-g(A) \in [0,1]~~~~~~~~:~~~uncertainty   \\
  m(\overline{A})=0 
  \end{array}
\right.
\label{eq:confdef}
\end{equation}
\vspace{-0.2cm}

\begin{figure}
\centering
\includegraphics[width=5cm]{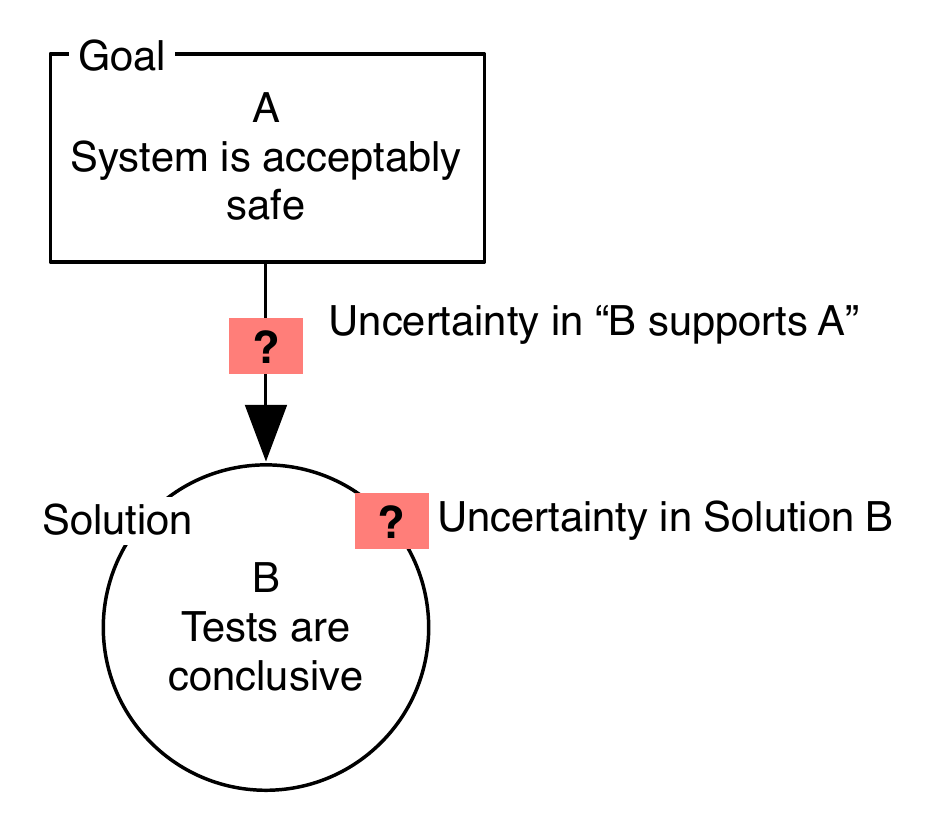}
\caption{Uncertainty points in a simple inference}
\label{fig:gsnconf}
\end{figure}

In the context of safety case, we consider two types of uncertainty sources, which are similar to those presented in \cite{HAW11} named "appropriateness" and "trustworthiness". For instance, in the very simple safety case presented in Figure~\ref{fig:gsnconf}, two sources of uncertainties may be identified:
\begin{itemize}
\item uncertainty in the fact that B is appropriate for the inference "A is Supported by B"
\item uncertainty in the solution B itself : are the tests trustworthy?
\end{itemize}

\section{Propagating Confidence}
\label{sec:propagconf} 
\subsection{Argument Types}

The very basic inference is the simplest one, "A is Supported by B". Nevertheless, most of arguments are more complex than direct one-to-one inference. For instance, let us consider the example presented with the main claim "A: System is fit for use", supported by both "B: Tests are conclusive" and "C: Formal verification has been performed". In that case, we can expect that both evidences independently increase the level of confidence in A. This concept is presented as "alternative argument" in \cite{CYR11}: even if there is no confidence in B, the fact that C also independently supports A will preserve some level of confidence. 

An another form of inference, is presented in the GSN «Hazard Avoidance Pattern» proposed in \cite{KEL97}, presented Figure~\ref{fig:gsn-example}. In that case, the main Goal "System is Safe", depends on all the sub goals together (we do not consider "Strategy" as a node, because it is only a descriptive element). Each of the premises covers a part of the goal. \cite{CYR11} propose to name such an argument a "complementary argument". 

\begin{figure}
\centering
\includegraphics[width=11cm]{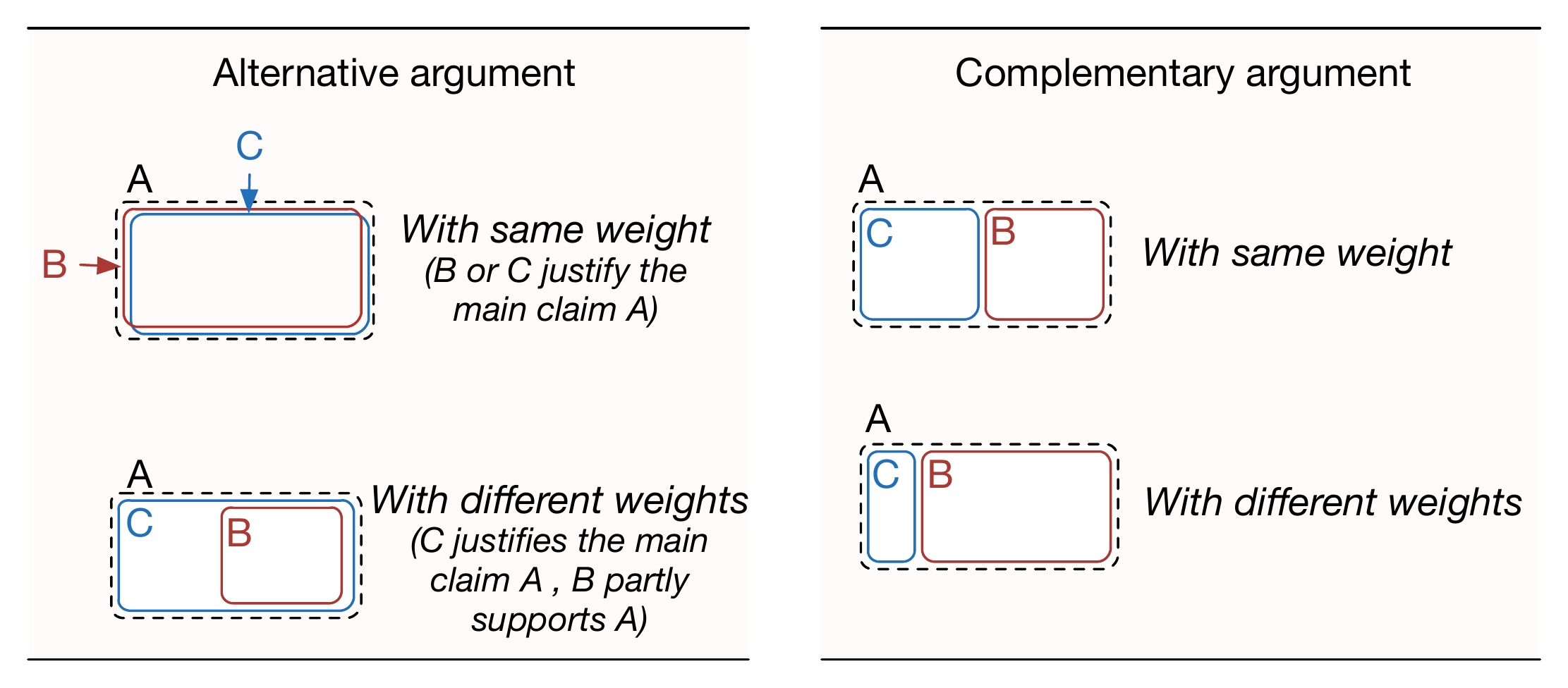}
\caption{Alternative and complementary arguments}
\label{fig:altcomp}
\vspace{-0.95cm}
\end{figure}

Figure~\ref{fig:altcomp} present those two types of arguments, with the inference "A supported by B and C". 
We also illustrate the fact that in both types of argument, the sub nodes may have a different weight in the overall confidence in the claim A. Other types of arguments may be included, as introduced in \cite{CYR11, ANA13}, but they are not included in this paper.

\subsection{Simple Argument}
\label{sec:argsimple}

\begin{figure}
\centering
\includegraphics[width=7cm]{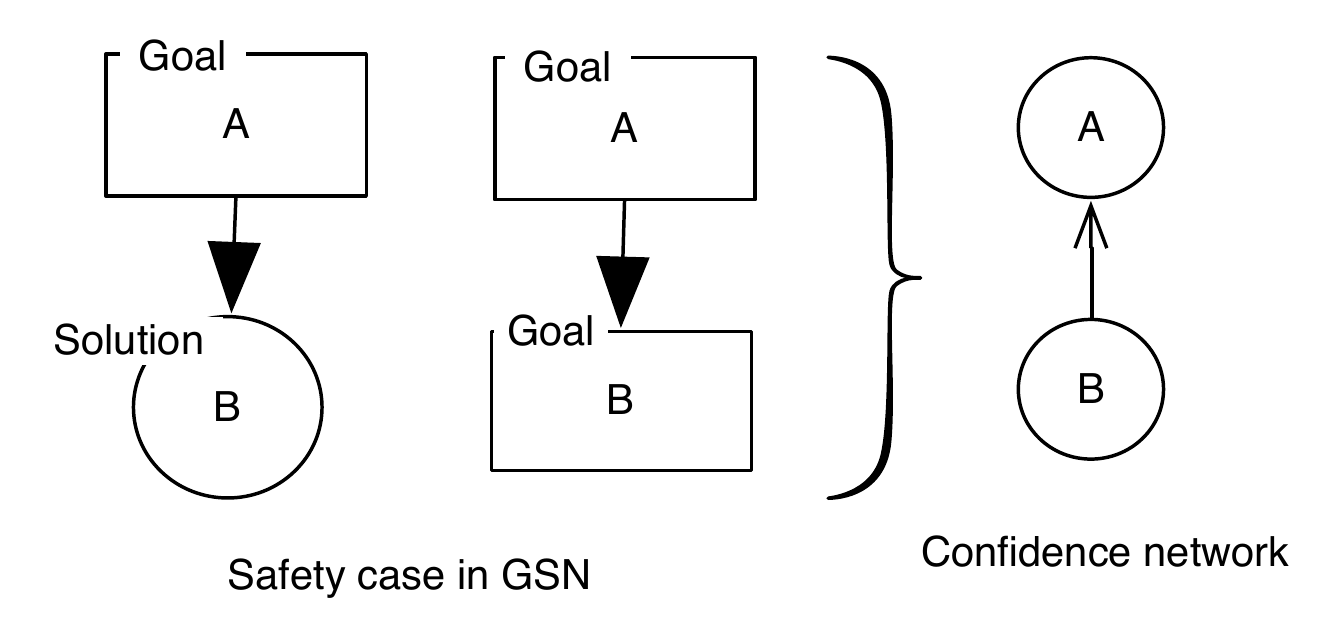}
\includegraphics[width=4cm]{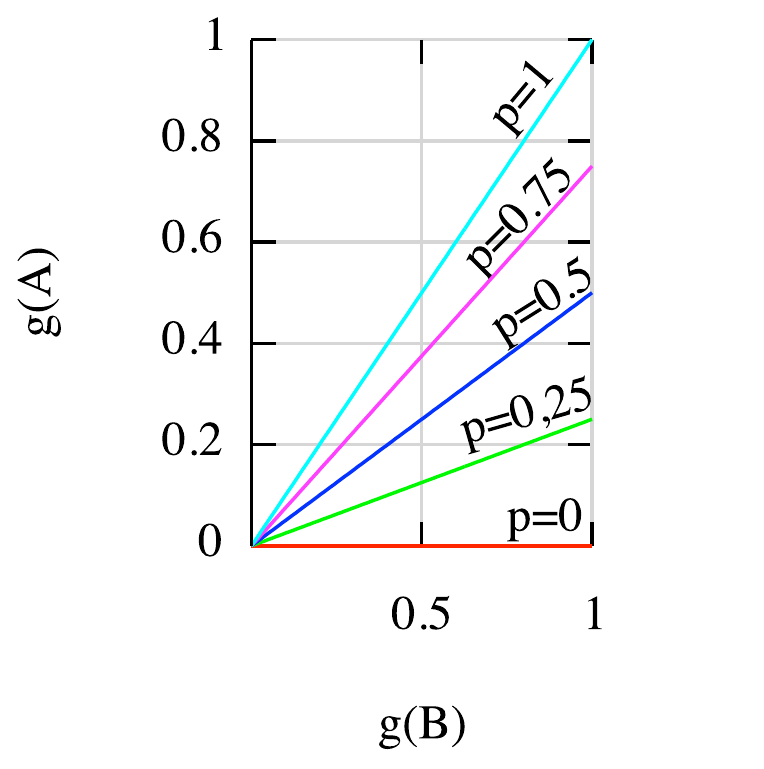}
\begin{tabularx}{\textwidth}{XX} \centering ~~~~~~~~~~~~~~~~~~~~~(a) &  \centering ~~~~~~~~~~~~~~(b)  \end{tabularx}
\caption{(a) GSN Simple argument transformation into confidence network and (b) $g(A)$ in function of $g(B)$, for $p\in [0;1]$}
\label{fig:infsimple}
\label{fig:infsimplegplt}
\vspace{-0.3cm}
\end{figure}

The basic inference, "A is supported by B" can apply to the cases (a) a goal is refined into a subgoal and (b) a goal is supported by an evidence, as presented in Figure~\ref{fig:infsimple}. 
In this case, the confidence network is represented like a BBN, using two nodes and one edge. We propose to use the following table to describe the confidence propagation:
\renewcommand{\arraystretch}{1.1} 
{
\begin{center}
\begin{tabular}{c||c|c}
g(B) & 0 & 1 \\ \hline g(A) & ~~~~0~~~~ & ~~~~~p~~~~
\end{tabular}
\end{center}
}
In this table, the confidence in A is estimated when there is no confidence in B (i.e. when $g(B)=0$), then $g(A)=0$, and when there is a maximum confidence in $g(B)$. In this case, the confidence in A depends on a factor $p$, which represents the confidence in the inference "A is supported by B". The final confidence 
is obtained using this table as a probability table: $g(A)=p*g(B)$. The result is a linear dependency $g(A)$, illustrated in Figure~\ref{fig:infsimplegplt} considering different values for p and g(B).

\subsection{Alternative Arguments}
\label{sec:argalt}
\begin{figure}[ht]
\centering
\includegraphics[width=10cm]{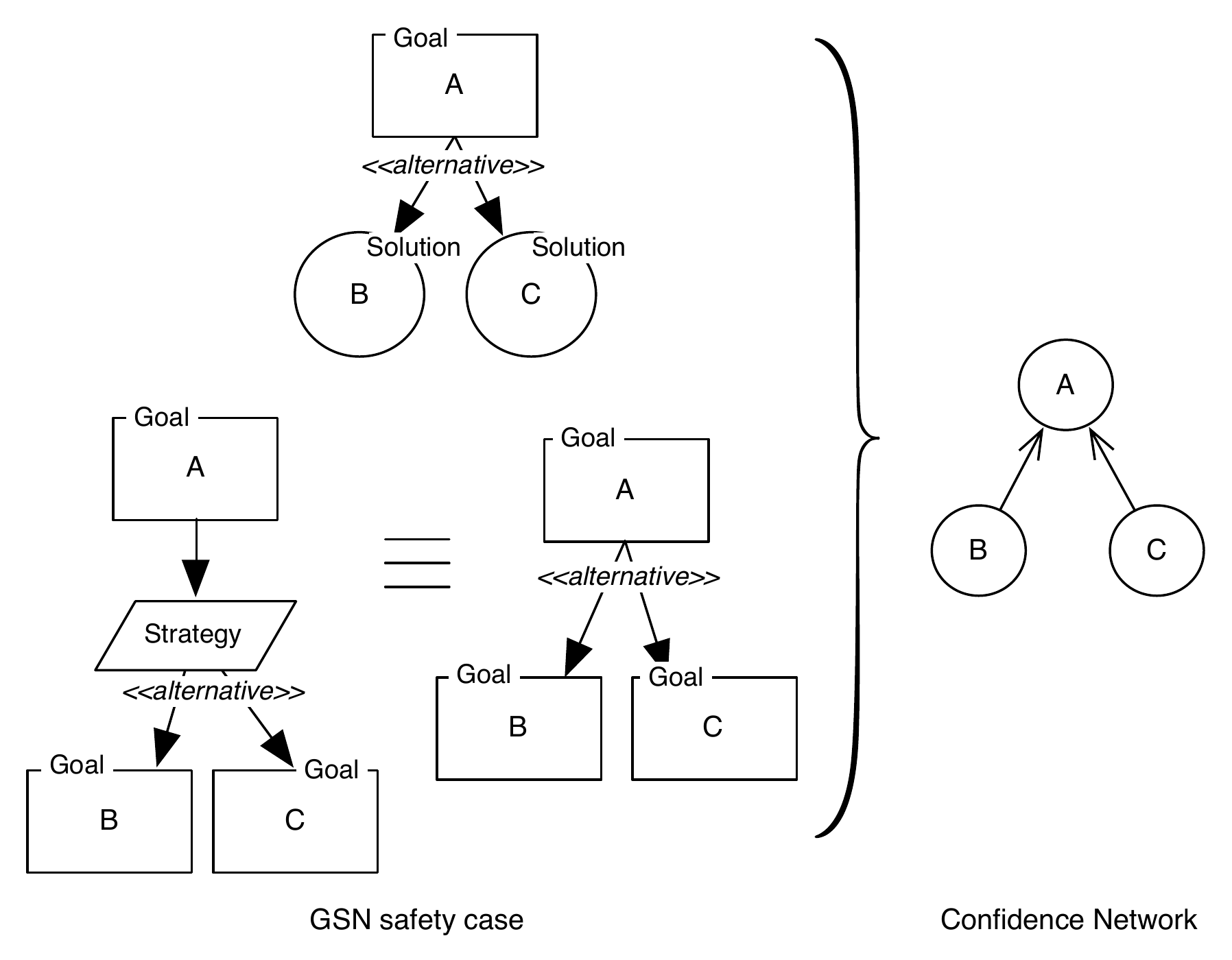}
\caption{Alternative argumentation transformation into confidence network}
\label{fig:mult}
\vspace{-0.55cm}
\end{figure}

As presented Figure~\ref{fig:altcomp}, several arguments may support a claim with an independent influence. It is important to note that in this Figure, we do not represent the confidence, but the way each argument supports the main claim. In this case, the confidence in A, may be increased by the confidence in both B and C. Such approach could be applied to Solutions or sub-goals as presented Figure~\ref{fig:mult}.
The Strategy node is not part of the confidence network, as it only gives explanations on the choices made for argumentation. 

We chose for this argument type to use a \emph{leaky noisy-or} as defined in probability theory \cite{DIE07}. It was originally introduced in \cite{PEA88}, and it is based on a logical OR between parent nodes ($Y_i$) and a child node($X$), but it includes the fact that the relationship between parents and the child node are not necessary deterministic. The $leaky$ effect corresponds to the fact that even when both parents (B and C) have 0-value probability, there is still a "leaky" probability for the child node. For probabilities, the mathematical function is, with $Y_v$ the set of $Y_i$ in state $\{True\}$:
\begin{equation}
P(X=\{True\} | Y_i)=1-(1-l) * \prod_{Y_i \in Y_v} (1-p_i)
\label{eq:leakynoisyor}
\end{equation}
with $p_i=P(X|Y_i,\{\overline{Y_i}\}_{j \neq i})$.
In its application to confidence, we do not consider the leaky effect, it is indeed obvious that if there is no confidence in B and C ($g(B)=g(C)=0$), then the confidence in A is zero, i.e. $g(A)=0$. Consequently, we obtain the following equation:
\begin{equation}
g(X | Y_i)=1- \prod_{Y_i \in Y_v} (1-p_i)
\label{eq:leakynoisyorg}
\end{equation}
According to \ref{eq:leakynoisyorg}, the resulting table for two parents is:
{
\begin{center}
\begin{tabular}{c||c|c|c|c}
g(B) ~& \multicolumn{2}{c|}{0} & \multicolumn{2}{c}{1}
\\ \hline 
g(C) ~ & 0 & 1 & 0 & 1
\\ \hline 
g(A) ~ & ~~0~~ & ~~q~~ & ~~p~~ & 1-(1-p)(1-q)
\end{tabular}
\end{center}
}

\begin{figure}
\vspace{-0.5cm}
\centering
\includegraphics[width=10cm]{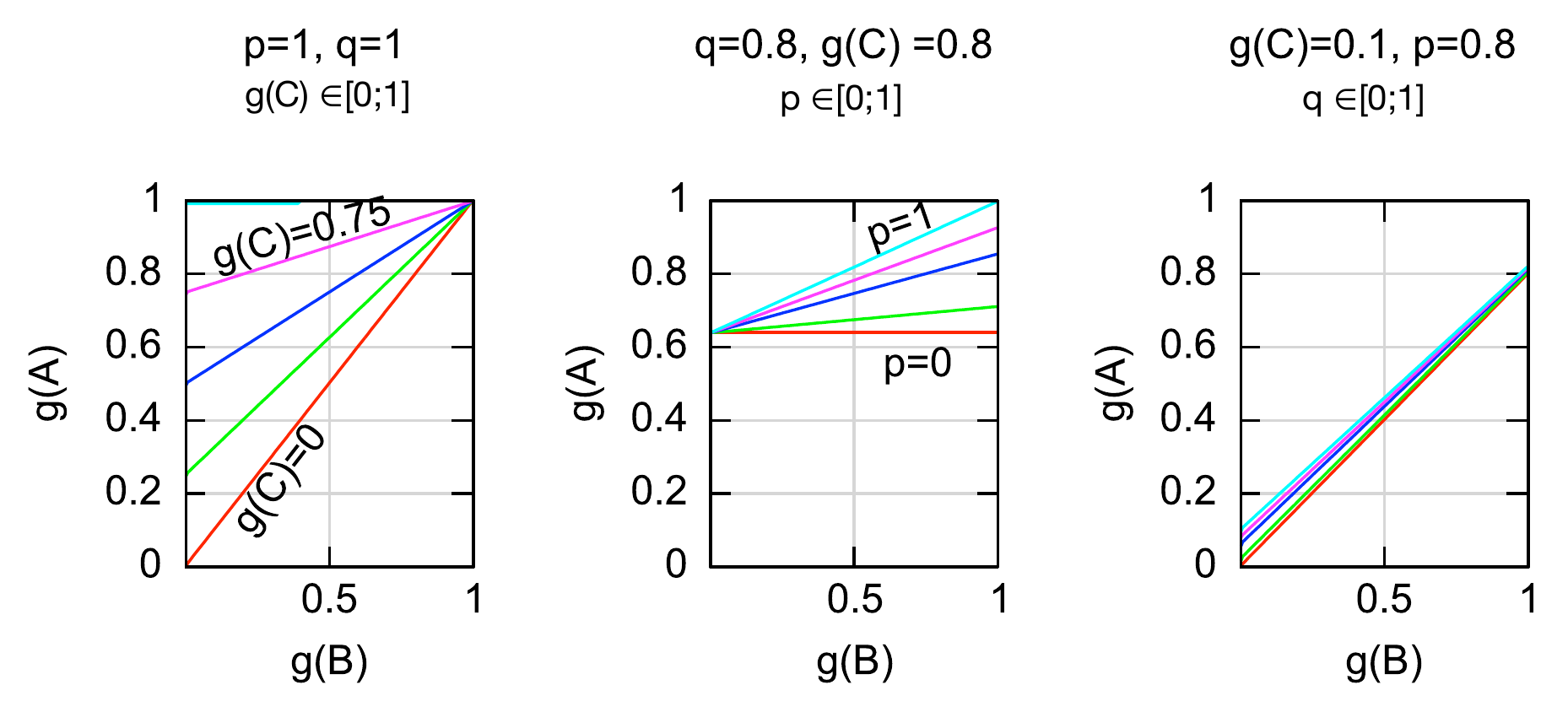}
\begin{tabularx}{\textwidth}{XXX} \centering ~~~~~~~~~~~~~~~~~~~~~(a) &  \centering ~~~~~~~~~~~~~~(b) & \centering ~~~~~~~~~~(c) \end{tabularx}
\caption{Alternative argument:  g(A), in function of g(B), g(C), p and q}
\label{fig:altgplt}
\vspace{-0.5cm}
\end{figure}

This leads to the confidence formula $g(A)=p*g(B)+q*g(C)-g(B)*g(C)*p*q$. $p$ and $q$ respectively represent the confidence in A in case one only has confidence in B or C. Figure~\ref{fig:altgplt} illustrates the evolution of confidence g(A) for different situations:
\begin{itemize}
	\item Figure (a) where $p=q=1$ illustrates that increasing the confidence in $g(B)$ alone or $g(C)$ alone, automatically increases $g(A)$.  For instance, for $g(C)=0.75$ and $g(B)=0.5$, the resulting confidence is 0.875. Confidence of 1 for A, occurs only if $g(B)$ or $g(C)$ reaches 1. 
	\item Figure (b) shows influence of p on $g(A)$. For a low confidence $p$ in the inference "A is supported by B", the confidence in A only depends on confidence in C ($g(A)$ is constant for $p=0$). 
	\item Figure (c) shows that for a low value of $g(C)$ (0.1), the variation of $q$, which is the confidence in the inference "A supported by C", has no effect on $g(A)$.
\end{itemize}

\vspace{-0.5cm}
\subsection{Complementary Arguments}
\label{sec:argcomp}

Complementary arguments are used when a set of solutions or subgoals are required simultaneously for supporting the main goal. However, a weight for each element is assigned to rate its relative importance. For instance, in the "Hazard Avoidance Pattern", some hazards may have a less impact on the overall safety, and the lack of confidence in their treatment, may induce less reduction in the main confidence, than for other more severe hazards. Several models are used in the literature for such arguments, such as simple And-gate \cite{Zhao12}, weighted mean \cite{DEN11}, or Noisy-And \cite{HOB12}. In our case, after several simulations, we decided to define our own Noisy-And, to obtain the trends that are relevant for complementary argumentation. 
In this case, we based our calculation on the uncertainty as defined in equation \ref{eq:confdef} and using the leaky noisy-or defined in equation~\ref{eq:leakynoisyor}, but taking for the leak $v=1-l$. We then obtain the following confidence table:

{
\begin{center}
\begin{tabular}{c||c|c|c|c}
$m(B,\overline{B}) $& \multicolumn{2}{c|}{0} & \multicolumn{2}{c}{1}
\\ \hline 
$m(C,\overline{C})$ & 0 & 1 & 0 & 1
\\ \hline 
$m(A,\overline{A})$& $ ~~1-v~~$ & $~~1-v.(1-q)~~$ & $~~1-v.(1-p)~~$ & $~1-v.(1-p).(1-q)~~$
\end{tabular}
\end{center}
}
To calculate the confidence table, we apply the relation $g(X)=1-m(X,\overline{X})$, and we also decided to fix $g(A)=0$ when $g(B)=g(A)=0$ (which should be obtain for whatever weight of B and C). We thus obtain the following table:
{
\begin{center}
\begin{tabular}{c||c|c|c|c}
$g(B) $& \multicolumn{2}{c|}{1} & \multicolumn{2}{c}{0}
\\ \hline 
$g(C)$ & 1 & 0 & 1 & 0
\\ \hline 
$g(A)$& $~~v~~$ & $~~v.(1-q)~~$ & $~~v.(1-p)~~$ & $~~~0~~~$
\end{tabular}
\end{center}
}
One main difference with other research works, lies in the interpretation of the parameters. In our case, $p$ and $q$ represent the weight of B and C to decrease confidence (increase uncertainty). In the context of confidence calculation, we also propose to introduce a relation between leak value $v$, $p$,and $q$ such as: $v=(p+q)/2$. Indeed, when p and q are lower than 1, it means that the confidence in the inference is less than one. The generalization of this constraint to a complementary argument with $n$ parents is:
\begin{equation}
v= \frac{1}{n} \sum_{i=1}^{n} p_i
\label{eq:contrainte}
\end{equation}
The values in the confidence table are:
\begin{equation*}
g(X| \overline{Y_1},...,\overline{Y_k})= v . \prod_{i=1}^{k}(1-pi)   
\end{equation*}
where $p_i$ represent the weight of $Y_i$ in the argument. 
\begin{figure}
\centering
\includegraphics[width=11cm]{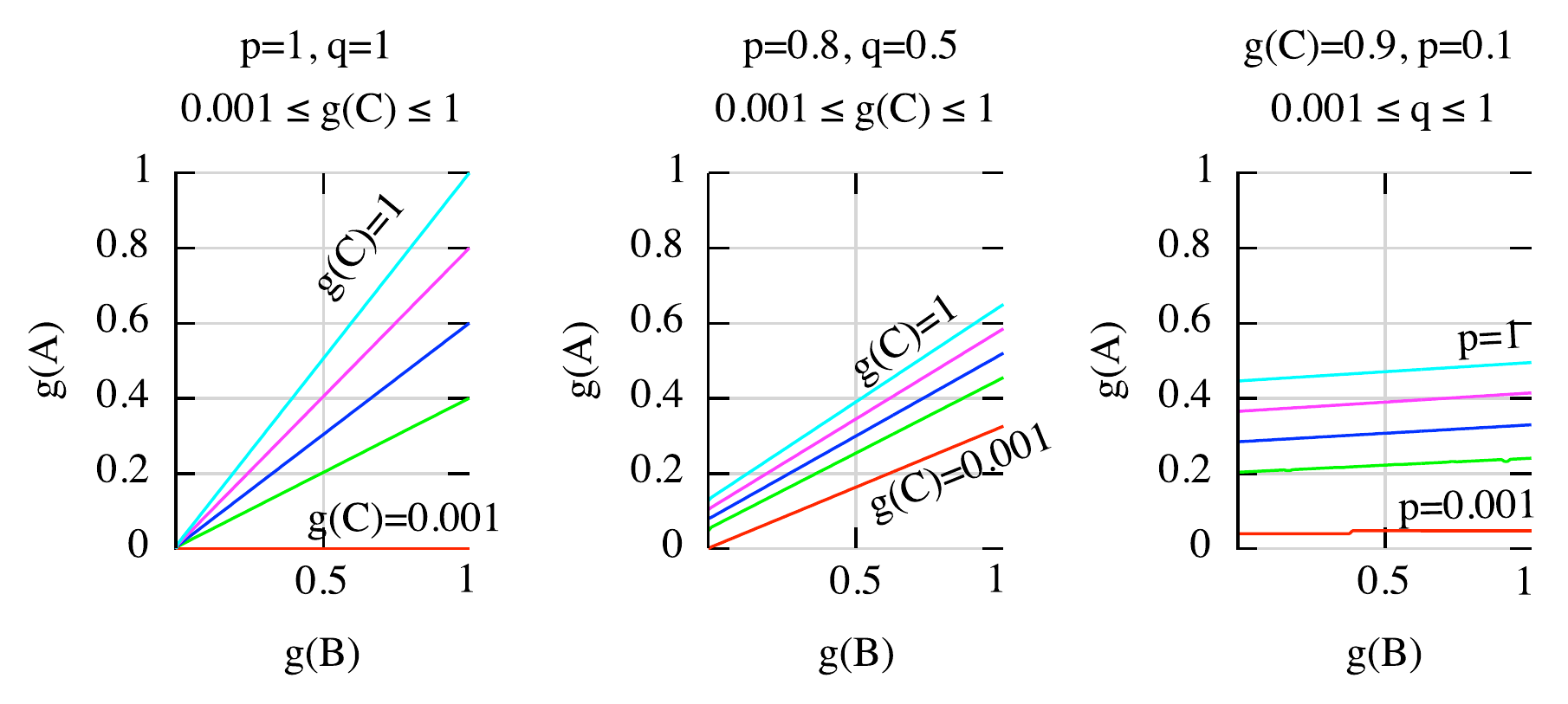}
\begin{tabularx}{\textwidth}{XXX} \centering ~~~~~~(a) &  \centering ~~~~~~~~(b) & \centering ~~~~~~~~~(c) \end{tabularx}
\caption{Complementary argument: g(A) in function of g(B), p and q}
\label{fig:infcomp1}
\vspace{-0.5cm}
\end{figure}
We consider in the following discussion that having a value of 0, for any confidence is not considered, has such an element (no confidence at all), will be removed from a safety argument. Figure~\ref{fig:infcomp1} presents the result for 2 parents, B, and C, and one child, A. In (a) and (b) we illustrate that when q decreases (q=1, q=0.5) then the influence of $g(C)$ decreases. On the figure, the lines for different values of $g(C)$ are close  depending only on $g(B)$ (with a value of 0.5, not presented here due to limitation space). We also illustrate in (b), that when p and q are less than 1, we obtain a residual confidence when $g(B)=0$ and $g(C)>0$. This is actually an expected result, because, when the weights are less than one, this means that the argument is not a perfect AND gate. In (c), p is low (0.1), which is interpreted as a low influence of $g(B)$, and characterized by the fact that all lines are nearly horizontals (i.e. with no influence of $g(B)$). A complete analysis of limits, which is not presented here, has demonstrated that the variations of $g(A)$ are compliant with a complementary argument~\cite{DOH15}.

\vspace{-0.3cm}
\subsection{Mixed Arguments}
\begin{figure}
\centering
\includegraphics[width=10cm]{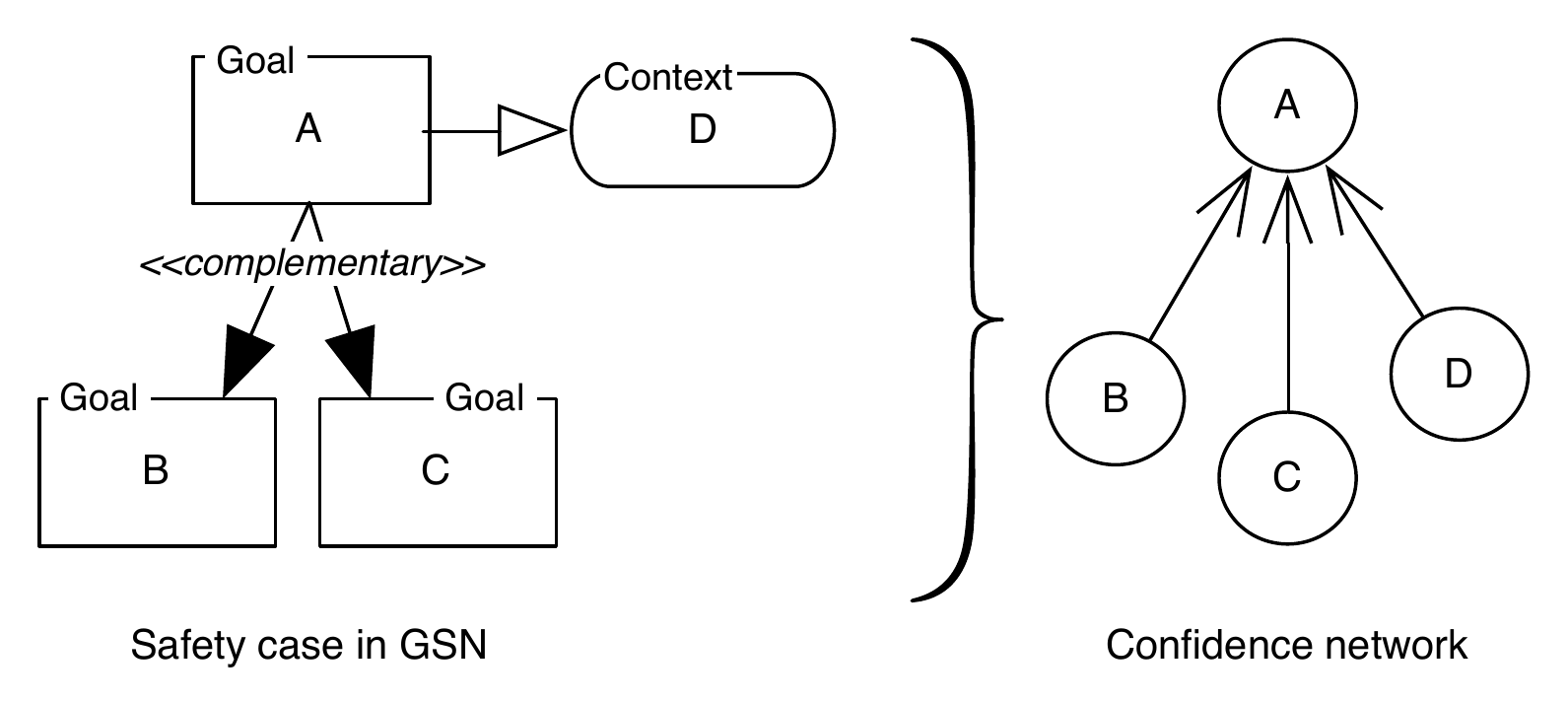}
\vspace{-0.2cm}
\caption{Mixed argumentation 1}
\label{fig:mult2}
\vspace{-0.75cm}
\end{figure}

The previous arguments could be used also to integrate the confidence in the GSN "Context" element. Indeed, a context is actually a complementary element for the considered argument. Figure~\ref{fig:mult2} presents a complementary argument, where a context has also been defined. In this case, the resulting network, is a node A, with three parents (B, C, D), and a noisy-and table for node A. When an alternative argument is used between B and C, then, an intermediate node I\_BC is included, with an alternative table for B and C. The confidence table in A is a noisy-and between D and I\_BC. 

\vspace{-0.3cm}

\subsection{Sensitivity Analysis}

We propose to perform a sensitivity analysis using a tornado graph. It is a simple statistical tool, which shows the positive or negative influence of basic elements on main function. Basically, considering a function $f(x_1,...x_n)$, where values $X_1, ..., X_n$ of the variables $x_i$ have been estimated, the tornado analysis consists in the estimation for each $x_i\in [X_{min},X_{max}]$, of the values $f(X_1,..., X_{i-1}, X_{min},X_{i+1},...X_n)$ and $f(X_1,..., X_{i-1}, X_{max},X_{i+1},...X_n)$, where $X_{min}$ and $X_{max}$ the maximum and minimum admissible values of variables $x_i$. Hence for each  $x_i$, we get an interval of possible variations of function $f$. The tornado graph is a visual presentation with ordered intervals. In our case, we estimate the intervals of $g()$ with $X_{min}=0$ and $X_{max}=1$.

\begin{figure}
\centering
\includegraphics[width=5.5cm]{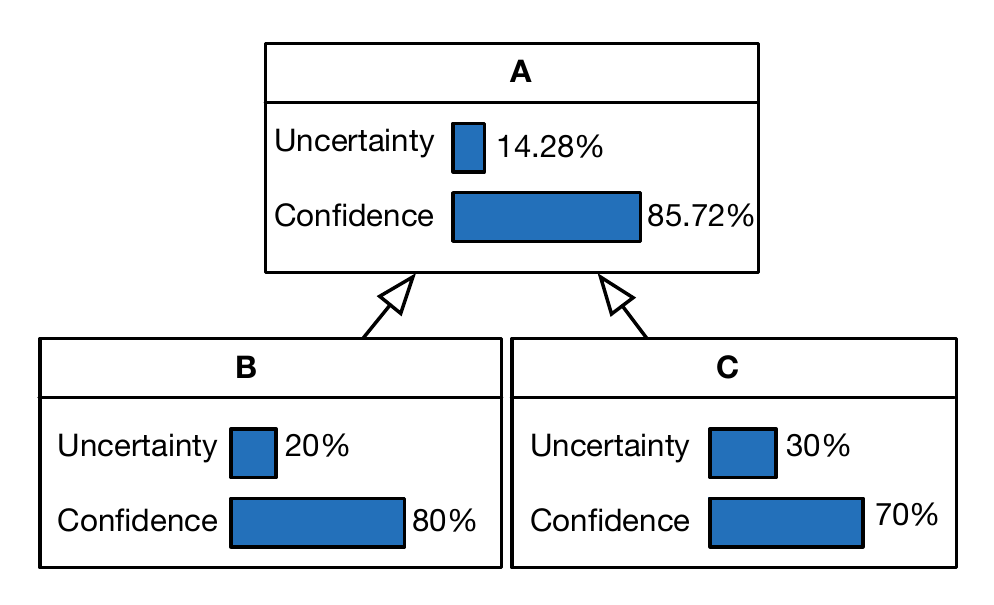}
\includegraphics[width=6.5cm]{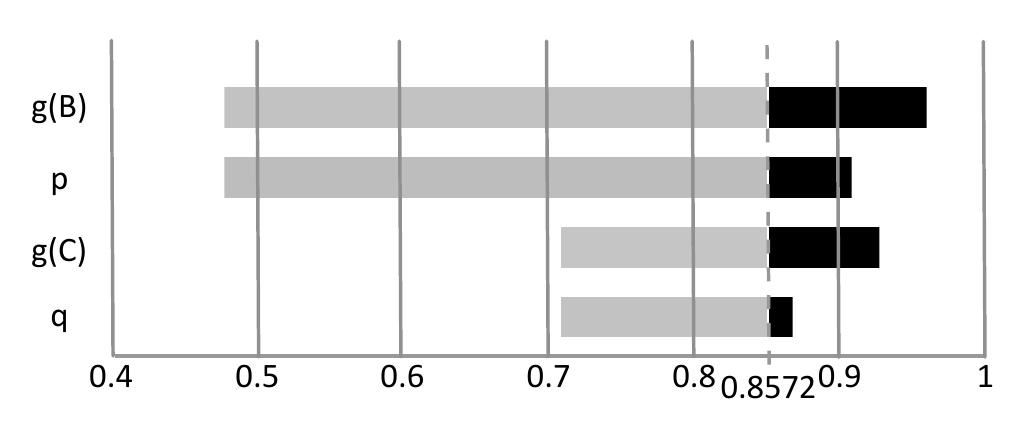}
\caption{(a) Example of an alternative argument with the tool Agenarisk and (b) Corresponding Tornado graph }
\label{fig:reseauagena}
\vspace{-0.5cm}
\end{figure}

If we take the example of alternative argument, with arbitrary values $q=0.7$ and $p=0.9$, we get the following table:
{
\begin{center}
\begin{tabular}{c||c|c|c|c}
g(B) & \multicolumn{2}{c|}{0} & \multicolumn{2}{c}{1}
\\ \hline 
g(C) & 0 & 1 & 0 & 1
\\ \hline 
g(A) & ~~0~~ & ~~0.7~~ & ~~0.9~~ & ~~0.97~~
\end{tabular}
\end{center}
}
If we choose the values of $g(B)=0.8$ and $g(C)=0.7$, the confidence table leads to the value $g(A)=0.8572$, also computed with the tool AgenaRisk\footnote{http://www.agenarisk.com}, presented Figure~\ref{fig:reseauagena}.
In this example, to determine the sensitivity to $g(B)$, we keep all the values for p, q, and $g(C)$, and only calculate the values $g(A)$ for $g(B)=0$ and $g(B)=1$ (we obtain the values 0.49 and 0.949).

The same approach is used for other variables p, q, and $g(C)$. The result is presented Figure~\ref{fig:reseauagena} (b).
In this tornado graph, g(B) appears to be the most influent parameter to decrease or increase the confidence in A. The left part is between 0.49 and 0.872, which means that if g(B) is equal to its lower limit, then the confidence in A could be reduced to 0.49. On the opposite, with a maximum value of g(B), then confidence in A could reach 0.949. 

Such an analysis leads to identify some sensitive points in a confidence network. This could be used to increase the confidence focusing first on the most positive sensitive points, or to focus on the elements where confidence should never be decreased (to consolidate the safety case confidence). Nevertheless, two main limits appear: it is not possible to identify combination of confidence variations, and such a diagram does not identify which variables are the easiest to increase. For instance, even if $g(C)$ appears to be less influent,  it may be easier to increase its confidence than the one in $g(B)$. Our approach does not focus on those aspects, but they are important points to study.

\section{Conclusion}
\vspace{-0.1cm}
This paper proposed a new approach for the definition and estimation of confidence in a safety case. We argue that it is important to have a separation between the safety case and the confidence case. Our aim is to analyze uncertainties that may be present in a safety case, using a sensitivity analysis. Our approach is based on the Dempster-Shafer theory for the definitions of confidence and uncertainty. But the constraint  $m(X, \overline{X})=0$, brings the main benefit of letting use mathematical tools, such as BBN. Hence, we proposed for most common safety case models in GSN, some transformation rules into a confidence network. We particularly introduce the use of noisy-or for alternative arguments, and an adapted version of noisy-and for complementary arguments. An experiment on a real case study of a rehabilitation robot \cite{GUI13} has been carried out \cite{DOH15}. A confidence graph of 65 nodes has been identified with only two alternative arguments (all the others were complementary). The complete analysis is still under development but, we were able to compute the complete graph and get a tornado graph in few minutes with the AgenaRisk tool with consistent results. 
In this paper, we only focus on the feasibility of a quantitative estimation of confidence, and its propagation in a confidence network. But this is obviously completely dependent on the determination of the confidence values themselves. As already mentioned, this important issue is not addressed in this paper, but is part of our future work. 

\vspace{-0.3cm}

\bibliographystyle{splncs03}
\bibliography{biblio}

\end{document}